\documentclass[11pt]{article}
\usepackage{geometry}
\usepackage{coling2020}
\usepackage{times}
\usepackage{url}
\usepackage{latexsym}
\usepackage{microtype}
\usepackage{graphicx}
\usepackage{placeins}
\colingfinalcopy 

\title{HinglishNLP: Fine-tuned Language Models for Hinglish Sentiment Detection}

\author{Meghana Bhange \\
  Verloop.io \\
  Bengaluru \\
  {\tt meghana@verloop.io} \\\And
  Nirant Kasliwal\\
  Verloop.io\\
  Bengaluru \\
  {\tt hi@nirantk.com} \\}

\date{}

\begin{document}
\maketitle
\blfootnote{

    \hspace{-0.65cm}  
    This work is licensed under a Creative Commons 
    Attribution 4.0 International Licence.
    Licence details:
    \url{http://creativecommons.org/licenses/by/4.0/}.
    %
    %
}
\begin{abstract}
Sentiment analysis for code-mixed social media text continues to be an under-explored area. This work adds two common approaches: fine-tuning large transformer models and sample efficient methods like ULMFiT \cite{ulmfit}. Prior work demonstrates the efficacy of classical ML methods for polarity detection. Fine-tuned general-purpose language representation models, such as those of the BERT family are benchmarked along with classical machine learning and ensemble methods. We show that NB-SVM beats RoBERTa by 6.2\% (relative) F1. The best performing model is a majority-vote ensemble which achieves an F1 of 0.707. The leaderboard submission was made under the codalab username \texttt{nirantk}, with F1 of 0.689.
\end{abstract}

\section{Introduction}
Code-mixing or code-switching refers to the use of two or more languages or speech variants together \cite{code_mixing}. This is commonly observed in informal conversations, especially those on social media, e.g. Twitter \cite{rudra-etal-2016-understanding} \cite{rijhwani-etal-2017-estimating}. While a small body of work does exist on code-mixing detection, this task focuses on polarity detection (Sentiment Analysis). We demonstrate the impressive performance of transfer learning for the task of sentiment detection in code-mixed context.  We discuss limitations of existing deep learning pre-trained models which are trained on "monolingual" text -- which has sentences in only one language. 

In this work, we demonstrate how it is beneficial to fine-tune the language model (LM) for a code-mixed setting like Hinglish, when Hindi is written in the Roman script. The labelled data for the sentiment classifier is from the task paper --- \cite{sentimix}. It consists of a total of 17000 tweets. The sentiment labels are positive, negative, or neutral, and the code-mixed languages are English-Hindi. The train data after the split has 14k tweets. The validation and test data contain 3k tweets each.

\section{Data}

We used two primary datasets. The data used for fine-tuning LM is from Twitter stream. The tweet stream data was used which contains {\raise.17ex\hbox{$\scriptstyle\mathtt{\sim}$}}1.9 million Hinglish tweets. We manually sampled 3k tweets to verify what fraction of them are Hinglish. This dataset consists of  {\raise.17ex\hbox{$\scriptstyle\mathtt{\sim}$}}86.5\% Hinglish tweets.  The 17k tweet Sentimix data \cite{sentimix} was used to further fine-tune the sentiment classifier. The Sentimix data contains 14,000 tagged tweets marked as positive, negative and neutral for training, and 3,000 for development and testing each. 

\subsection{Mining and Filtering Twitter Corpus}

Pre-training LM require large datasets. In order to enable this, we gathered {\raise.17ex\hbox{$\scriptstyle\mathtt{\sim}$}}1.9 Million tweets from the Twitter 1\% sample stream for the entire year of 2018. We curated a seed dictionary of Hinglish words and their spelling variants. Next, we calculate the Jaccard Index \cite{doi:10.1111/j.1469-8137.1912.tb05611.x} between our seed dictionary and every tweet. For values above 0.6, we mark the tweet as ``Hinglish tweet". This threshold value 0.6 was selected empirically, by evaluating Jaccard values for 200 tweets.

Next, every token which is present in the tweet, but missing in our dictionary is marked as a Candidate token $C_t$. We remove duplicates to get our set of unique candidate tokens $C$. 
For every unique $C_t$ in $C$, we manually review and add to our dictionary. A known limitation of this iterative-expansion dictionary based approach is that it starts out with a bias for smaller tweets with fewer total tokens. Hence, we repeated this exercise in batches of 10,000 tweets each -- till we saw two batches of full 280 character tweets. We marked these tweets as ``highly likely" to be Hinglish. These were roughly 160,000 tweets. A secondary split of roughly 380,000 tweets was marked as ``possibly" Hinglish. Both of these were primarily used for training or fine-tuning the LM backbone. 

The {\raise.17ex\hbox{$\scriptstyle\mathtt{\sim}$}}1.9M tweets are composed of 3 different splits, with differing Hinglish percentage. The first split of 162K tweets is enriched to 86.5\% Hinglish, with 13\% tweets being empty, or non-Hinglish in other ways. The second split of 384K tweets is enriched to 89.6\%. The remaining \~ 1.4M tweets are expected to have 83\% Hinglish tweets. We randomly pulled 1K tweets from each of these splits to get these estimates. The estimate is hence, prone to sampling biases/errors. To re-iterate, this added dump is neither tagged with polarity nor pure Hinglish.

\subsection{Release}
The dataset is released on Github.\footnote{https://github.com/NirantK/Hinglish} Since Twitter discourages releasing the text directly, tweet\_ids are shared. This leaves the user to pull the specific tweets using Twitter's Developer API. 

\subsection{Cleaning and Pre-processing}
We de-duplicated the 1.9M tweets corpus using a string equality check. We also de-duplicated tweets using the meta-information in the JSON when a ``retweeted" text is included twice. The text was pre-processed before data were introduced to the model. The pre-processing included removal of both external links and shortened twitter links. The ``@" was replaced with with ``mention". Similarly, ``\#" was replaced with the word ``hashtag". Emojis were converted to text equivalent using the emoji package \cite{emoji}. During this stage, both the datasets (SemEval and Twitter Large Supervised Dataset) are pre-processed with identical code, both during training/fine-tuning and inference.s

\section{Experiments}
\subsection{Training and Fine-tuning}

The Language Models (LMs) were fine-tuned on the entire Twitter Stream Sample. We used held out about 10\% for measuring LM perplexity. Classifiers were trained using 14000 tweets from the 17000 tweets in the SemEval training corpus. The linear layers were fine-tuned on the SemEval training corpus for 3 epochs for all experiments. The fine-tuning parameters for the BERT-family sentiment classifiers are referenced in Table \ref{bert-class-config}. For Attention dropout and hidden dropout, the parameters were empirically chosen using random grid-search with a range of 0.1 to 0.9. The range considered for Adam Epsilon was 1e-8 to 9e-8 with 1e-8 granularity. Learning rates varied from 1e-7 to 1e-4. These parameters were combined with two learning rate schedulers, a linear learning rate scheduler and a cosine learning rate scheduler. The training for models which took place in two steps: First, the pre-trained language model was fine-tuned using the 1.9M tweets. Second, this fine-tuned deep LM was used as an encoder for training the polarity classifier using the 14K tagged tweets from Sentimix. 

\begin{table}[t]
\centering
\begin{tabular}{lllll}
\hline \textbf{Parameter} & \textbf{BERT }&\textbf{Hinglish}&\textbf{RoBERTa}&\textbf{DistilBERT}  \\
 & \textbf{Multilingual}&\textbf{Fine-tuned BERT}&&\\ \hline
Attention Dropout Probability &0.4 & 0.4&0.1&0.6\\
Hidden Dropout Probability & 0.3 & 0.3&0.1&0.6\\
Adam Epsilon & 3e-8 & 1e-8&5e-8&1e-8\\
Warmup Steps &100 & 100&0&100\\
Maximum Learning Rate & 5e-7 & 5e-7 & 4e-5 &3e-5\\
Learning Rate Scheduler & linear & linear&linear&cosine\\
\hline
\end{tabular}
\caption{\label{bert-class-config}Fine-tuning Parameters for BERT Family Classifiers}
\end{table}

\subsection{Evaluation}
During the competition, we used multiple methods of evaluation and different train-test splits. In this work, the \textbf{test dataset} refers to the officially released test set of 3,000 tweets. The performance numbers have been updated to reflect the same. We chose to ignore the validation set from SemEval for evaluation because most of our LMs had consistently very high performance of 0.95 F1 or more on the set. The F1 score used for internal evaluation is macro-F1 while the leaderboard submission uses weighted-F1. The F1 scores that are shown in the result table are from internal evaluation and thus are macro-f1. 


\section{Modeling Approaches}
\subsection{NB-SVM}
NBSVM is the approach proposed by \newcite{NBSVM}, which performs well on text classification in tasks like sentiment classification. It takes a linear model such as SVM (or logistic regression) and incorporates the possibilities of Bayesian by replacing terms with Naive Bayes log-count ratios. The NBSVM implementation was borrowed as is from \newcite{NBSVMImplementation}. The motivation for using NBSVM is that they are comparatively faster to train as opposed to deep learning models. We chose $C=4$, the inverse regularization parameter.

\subsection{ULMFiT: Universal Language Model Fine-tuning for Text Classification}
We used AWD-QRNN \cite{QRNN} instead of AWD-LSTM \cite{ulmfit} for pre-training and fine-tuning. We used Sentence Piece\cite{SentencePiece} for text tokenisation. The intent was to capture sub-word level features. Vocabulary Size of the sentencepiece tokenizer was 8000 and was trained on 540k tweets to save compute time. For the ULMFiT-QRNN model batch size of 1024 was used while training both the LM and classifier (linear) layers. AWD-LSTM gives an F1 0.48 on the test set where as AWD-QRNN performs with an F1 of 0.650. The hypothesis which could explain this is that tweet length, which is typically less than 140 characters, is too short for LSTM is learn a meaningful pattern.

\subsection{BERT Multilingual}
\textbf{BERT-base-multilingual-cased} \cite{BERT}, without any fine-tuning of LM on Hinglish data,  was used to train the sentiment classifier. It is trained on cased text in the top 104 languages with the largest Wikipedia corpora. The linear layers were trained/fine-tuned without updating the frozen backbone for 3 epochs.

\subsection{Hinglish Fine-tuned BERT}
The base model for fine-tuning BERT LM on Hinglish data was \textit{BERT-base-multilingual-cased} \cite{BERT}. Both the backbone and linear layers of the LM were fine-tuned. This was on a pre-processed Twitter Stream Sample (described in the previous section) over 26,000 iterations. It was trained for a total of 4 epochs.  Training batch size was four and vocab size 119,547. The perplexity of the fine-tuned LM was 8.2. The trained BERT tokenizer and model were utilized for fine-tuning classifier. 

\subsection{RoBERTa}
RoBERTa \cite{DBLP:journals/corr/abs-1907-11692} is a robustly optimized BERT pre-training approach. It is trained over longer sequences and removes the next sentence prediction task from BERT pre-training.  The base model for fine-tuning the LM-backbone for RoBERTa on Hinglish data was \textit{RoBERTa-base}. The LM was fine-tuned on a pre-processed unsupervised twitter dataset over 25,000 iterations. It was trained for a total of 3 epochs. Training batch size was four and vocab size 50265. The perplexity of this model was 7.54. 

\subsection{DistilBERT}
DistilBERT \cite{DistilBERT} uses the technique of knowledge distillation to improve the performance of BERT and create a smaller distilled version of the model. The LM-backbone was fine-tuned on a pre-processed unsupervised twitter dataset over 49,000 iterations. It was trained for a total of 6 epochs.  Training batch size was four and vocab size 28996. The perplexity of the fine-tuned LM-backbone for distilBERT was 6.51 and the base model used for fine-tuning the LM was \textit{distilbert-base-cased}.

\subsection{Ensemble \label{ensemble_title}}
\begin{figure}[t]
\includegraphics[width=15cm]{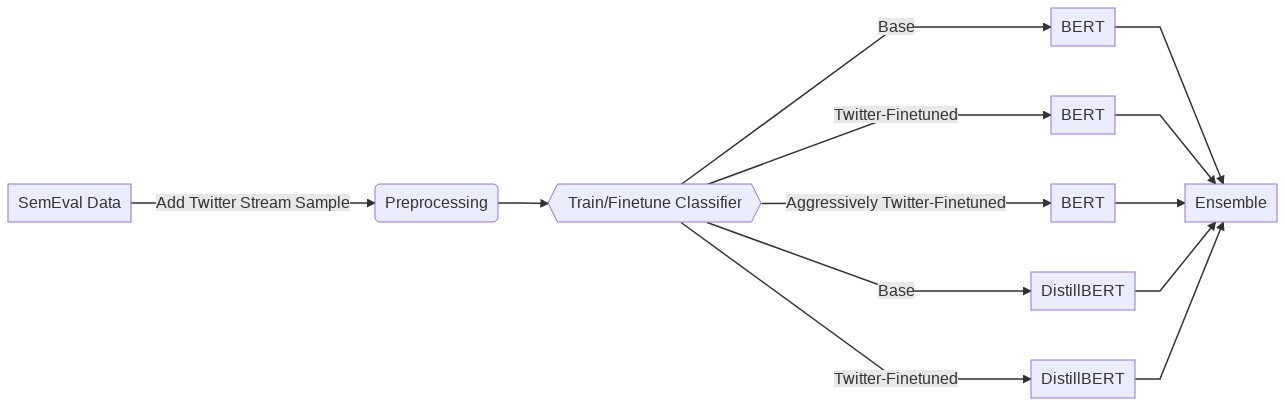}
\caption{\label{ensemble} Once pre-processed, the data is used for predicting results which are then passed to the ensemble described in section \ref{ensemble_title}.}
\end{figure}

For the final submissions, three variations of BERTs and two variations of DistilBERT were used. These were the top 5 selected based on their validation accuracy. For the ensemble, Weighted Majority Voting, by using the prediction confidence (0 to 1 scale) as the weight;  The ensemble methodology and its usage in our case is described in Figure \ref{ensemble}.

\section{Results}
The result for the experiments are summarized in Table \ref{class-perf}. Out of all the techniques used on test-data, Weighted majority vote ensemble with LR funneling gained a significant edge when it comes to F1 score. Traditional machine learning models like NB-SVM show a comparative performance. 

\begin{table*}[t]
\centering
\begin{tabular}{llllll}
\hline
\textbf{Model} & \textbf{Accuracy} & \textbf{Precision} & \textbf{Recall} & \textbf{F1}&\textbf{LM-Perplexity}\\
\hline
\textbf{Majority Vote (Ensemble)} &\textbf{0.704}&\textbf{0.709}&\textbf{0.707}&\textbf{0.707}&--\\
\textbf{DistilBERT-Base-cased} &0.685&0.690&0.685&0.687&\textbf{6.51}\\
\textbf{Logistic Regression Funnel (Ensemble)} &0.678&0.678&0.696&0.684&--\\
\textbf{BERT-Base-Multilingual-Cased}  & 0.680&0.692&0.678& 0.677&8.2\\
\textbf{NB-SVM (Ensemble)} &0.667&0.666&0.685& 0.673&--\\
\textbf{ULMFit AWD-QRNN} &0.645&0.646& 0.660&0.650&21.0\\
\textbf{RoBERTa-base} &0.630&0.629&0.644&0.635&7.54\\
\hline
\end{tabular}
\caption{\label{class-perf}
Results on sentiment classification where the F1 is performances of the model on test-data provided by Sentimix. The models with LM-backbones are provided with the perplexity of the fine-tuned LM where as the ones without are denoted by NA. 
}
\end{table*}

\section{Future Work}

There are three main incremental directions of improvements: data, methods, adopting techniques from text classification. For instance, initial tweet data had a lot of truncated tweets, using tweet\_ids to get an entire tweet would enrich our inputs. The training data can also be augmented in a wide variety of ways such as using vector similarity \cite{ma2019nlpaug}.

We can investigate other methods which might help in understanding missed case. Sentence embeddings for Hinglish, similar to InferSent \cite{Conneau2017SupervisedLO} or Universal Sentence Encoding \cite{USE} may be promising, in addition to Skip Thought or other sentence vectorisation methods, as well as exploring the performance of models which do not focus on transfer learning like R-CNN, and LSTMs. 

Lastly, a wide variety of deep learning tricks and methods could be used, such as label smoothing \cite{whylabelsmoothing}, which can help in generalising better beyond the small training sample.

\section{Conclusion}
We demonstrate that ensembles of classical Machine Learning models, even NB-SVM exhibit competitive performance and can in fact be better than some Transformer baselines. It is still worthwhile to implement simple classical baselines. 
Additionally, we hope that the released dataset and models \footnote{https://github.com/NirantK/Hinglish} will encourage readers to investigate this further. 
\FloatBarrier

\bibliographystyle{coling}
\bibliography{semeval2020}

\end{document}